# Algorithm for Semantic Network Generation from Texts of Low Resource Languages Such as Kiswahili

Barack Wamkaya Wanjawa*
University of Nairobi
Department of Computer Science
P.O. Box 30197
Nairobi 00100, Kenya
wanjawawb@gmail.com
*Corresponding author

Lawrence Muchemi
University of Nairobi
Department of Computer Science
P.O. Box 30197
Nairobi 00100, Kenya
lmuchemi@uonbi.ac.ke

Evans Miriti
University of Nairobi
Department of Computer Science
P.O. Box 30197
Nairobi 00100, Kenya
eamiriti@uonbi.ac.ke

**Abstract**
Processing low-resource languages, such as Kiswahili, using machine learning is difficult due to lack of adequate training data. However, such low-resource languages are still important for human communication and are already in daily use and users need practical machine processing tasks such as summarization, disambiguation and even question answering (QA). One method of processing such languages, while bypassing the need for training data, is the use semantic networks. Some low resource languages, such as Kiswahili, are of the subject-verb-object (SVO) structure, and similarly semantic networks are a triple of subject-predicate-object, hence SVO parts of speech tags can map into a semantic network triple. An algorithm to process raw natural language text and map it into a semantic network is therefore necessary and desirable in structuring low resource languages texts. This algorithm tested on the Kiswahili QA task with upto 78.6% exact match.

**Highlights**
- Languages, both low and high-resource are important for communication.
- Low resource languages lack vast data repositories necessary for machine learning.
- Use of language part of speech tags can create meaning from the language.
- An algorithm can create semantic networks out of the language parts of speech.
- The semantic network of the language can do practical tasks such as QA.

**Keywords**

## 1.0 Introduction

Low-resource resource languages, which includes many African languages, have not been widely used on the internet for practical user needs such as information retrieval, summarization, machine translation, disambiguation, question answering, or internet search. This is attributed to the low research focus on these languages due to lack of readily available tools to facilitate their processing. Any natural language, spoken by people, would usually need to be processed in some way for it to be useful for practical user applications on the computer systems or the web (King, 2015).

Structuring of natural language (NL) is a prerequisite step in the processing of the NL text for machines processing tasks. Structuring of high-resource languages has been possible and has largely been done using the many available processing tools and training datasets. However, low resource languages have not spurred much research interest, hence the available tools and datasets are comparatively few. This could be due to lack of initial tools to even process the language data where it is available, or lack of training data when the need for training of models is necessary (Hirschberg & Manning, 2015).

The processing of languages for use on the web or on other computer applications involves steps such as tokenizing, canonizing, normalizing to Unicode, stop work removal, synonym processing, stemming and even named entity recognition. Thereafter, a knowledge representation is created using either a probabilistic or embedding approach to represent the language. This is the representation that the computer processes to realize the practical user tasks such as searching, querying or information retrieval (Pennington et al., 2014). These representations require training data before they are useful in practical systems (Y. Li et al., 2016; Yan & Jin, 2017).

Knowledge representation (KR) refers to the use of symbols to represent propositions (Brachman & Levesque, 2004). NL, which is what humans use for communication, is not directly representable in KR used by machines. This is because NL suffers from ambiguity, inconsistency, and expressiveness, while machine learning agents need explicitly defined assertions to always represent correct meaning. NL is therefore difficult to directly process in computing systems without some KR modelling methods. Some of these modelling methods are frames, description logic, fuzzy logic or graphically (semantic networks, conceptual maps, and conceptual graphs).

However, not all KR requires data for training. One such KR being the graphical methods e.g. semantic networks (SN), concept maps and conceptual graphs. The concept of interconnectedness of data, as represented in semantic networks, is already exploited in linked data systems. Linking data enables data from diverse sources to be accessed through one entry point that in turns links to other data sources. Linked data, which is just a series of interrelated triples, each linking another, is the request of the semantic web, which enables the querying of information from the data network (Berners-Lee, 2006).

It is therefore possible to exploit the possibility and advantages of linking data to process knowledge and to represent it for other downstream applications. Therefore, NL text structured as a SN is already good enough to be used in applications that require structured text, such as Question Answering tasks. This is therefore done while bypassing the major bottleneck of processing low-resource languages, which is the lack of training data. The NL text is therefore ready, as is, for structuring into a machine understandable format, hence a method of resolving the problem of low interest in low-resource languages (Besacier et al., 2014). A step-by-step method of formatting this text, just based on the language structure itself is therefore desirable and can be represented as an algorithm.

The study of the structure of NL as used by humans, confirms that they have a particular format in the construction of sentences and derivation of meaning. The popular structures are of three formats, namely, subject-verb-object (SVO), subject-object-verb (SOV) and verb-subject-object (VSO) (Gell-Mann & Ruhlen, 2011; Marno et al., 2015). One such low-resource languages is Kiswahili, and it is an SVO language (Sánchez-Martínez et al., 2020). Kiswahili, also known as Swahili, is used by over 140million users worldwide and is predominantly spoken in the East African countries of Kenya and Tanzania as the national language. The language is also used in many different countries in the world such as Australia, Canada, Saudi Arabia, UK and USA (omniglot, 2021). It is therefore a language of international importance worthy of resourcing (Hirschberg & Manning, 2015).

Interestingly, the knowledge representation of a semantic network (SN) is a triple of subject-predicate-object (SPO), while some languages, such as Swahili, are structured as SVO. Careful study of the language structure (SVO) and SN structure (SPO) shows that the language can be mapped into a semantic network to give it meaning, through an SVO to SPO mapping. This is done at the language structure level (part of speech), without the need for training data, which is usually not available for low-resource languages. This means that a rule-based system is a candidate solution for structuring the language. Such rules can easily be structured into an algorithm to guide any computer processing system on how to do the SVO to SPO mapping.

There are many NL tasks that users derive from NL texts. One of the tasks, Question answering (QA), remains a difficult NL problem with ongoing active research (De Cao et al., 2019; Welbl et al., 2018; Yao et al., 2019). There have been different approaches employed in question answering tasks. The deep learning method of Graph Recurrent Network (GCN) using GloVe word embeddings method trained on Wikihop data can process natural English language text and then do QA task (Song et al., 2018). Other deep learning systems such as Embeddings from Language Model (ElMo) (Peters et al., 2018) and deep learning models such as BERT already process high-resource texts to a high degree of accuracy, provided they are trained with large amounts of data (The Stanford Question Answering Dataset, 2021). QA is therefore a candidate test case for confirming effectiveness of the SPO-to-SVO mapped knowledgebase.

The objective of this research is therefore to develop a rule-based algorithm that maps the SVO structure of a low-resource language, into the SPO structure of a SN to then give the language a structure. This structure is then exploited directly by a computer to understand, process, and do machine processing tasks on the language. As proof of concept, the developed SN from the low-resource language is tested on the NL task of question answering to gauge applicability. This research therefore benefits the many low-resource languages that are not being exploited now, mainly due to lack of training data.

The rest of the article is structured as follows: section 2 highlights the methodology of developing the algorithm proposed in this research, while section 3 shows the results of using the algorithm in typical applications. Sections 4 and 5 giving the discussions and conclusion of the research findings.

## 2.0    Methodology

This research aims at developing an algorithm to guide in the processing of raw text of a low-resource language, with the focus on the Kiswahili language. The output of the algorithm is a semantic network triple formulated from NL text, just by a review of the parts of speech tags of the text itself.

The algorithm developed in this research is based on a model of Swahili language processing from previous works (B. Wanjawa & Muchemi, 2021). The model provided the pipeline stages and highlighted the SVO identification stage as a critical processing step, as now expounded in this research. Fig. 1 below shows the flowchart that describes the key stages in typical SVO identification from NL text.

The flowchart shows that the initial processing stage after reading the input sentence is part of speech (POS) tagging of the text. This tagging is important since the modeling of the language relies on the POS tags and not the words themselves. It is the POS e.g. noun or verb that leads to the realization of a triple of subject-verb-object (SVO), which is then mapped into the semantic structure of SPO.

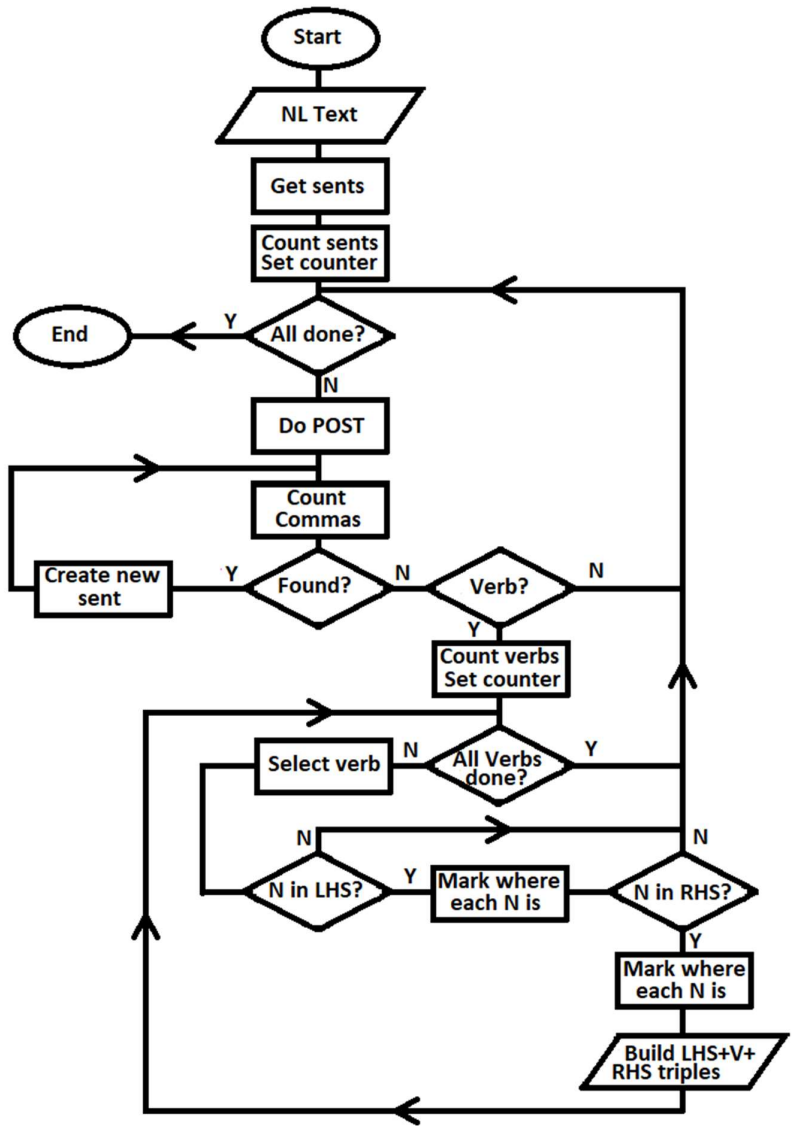

**Fig. 1.** Flowchart for semantic network generation from natural language text (Source: author)

The algorithm formulated in this research is shown in Fig. 2 below. It gives the step-by-step process of generating candidate SVO triples, and how to finally decide the most suitable SVOs for extraction as the final SVOs for the semantic network's SPOs.

```
1.  Start
2.  Read the NL text
3.  Count no. of sentences
4.  Set max_sentences
5.  Set counter = 0
6.  For sents from 1 to max_sent
7.      Read the sentence
8.      POST the sentence
9.      Count the COMMAs
10.     If COMMA found DO Procedure COMMA_found
11.     If NO COMMA found DO Procedure no_COMMA
```

```
12.         //Procedure no_COMMA
13.         If no COMMA in phrase:
14.             Count VERBs
15.             Set max_verbs
16.             If max_verbs > 0 DO Procedure VERBS_found
17.             //Procedure VERBs found
18.             verb_count = 0
19.             Repeat
20.               Set verb_count = verb_count + 1
21.               Read POST of LHS of V
22.               If N exists on LHS of V
23.                 Determine all Ns on LHS of V and create array of the Ns
24.                 Check RHS of V
25.                 If N exist on the RHS of V
26.                    Determine all Ns on RHS of V and create array of the Ns
27.                    // stitch each LHS token to every RHS token with V as connector
28.                    Set LHS_N_max
29.                    LHS_N_count = 0
30.                    Repeat
31.                        Set LHS_N_count = LHS_N_count + 1
32.                        New_LHS_phrase = LHS_N(count) + append the V
33.                        Set RHS_N_max
34.                        RHS_N_count = 0
35.                           Repeat
36.                           Set RHS_N_count = RHS_N_count + 1
37.                           New_triple = New_LHS_phrase + append the RHS_N(count)
38.                           Until RHS_N_count > RHS_N_max
39.                    Until LHS_N_count > LHS_N_max
40.                 else exit // the triple not possible, no N on RHS, despite having S-V
41.               else exit // the triple not possible, no N on LHS of V
42.             else DO procedure OTHER_Rules // triple not possible, no V, try other rules
43.             //Procedure OTHER_RULES – set any other rules e.g. 'is-a' rules
44.             If phrase has N-N – New_triple = N- <is_a> -N
45.             If phrase has N-NUM – New_triple = N- <is_a> -NUM
46.             etc.
47.             else exit // triples not possible with current POS structure

48.         //Procedure COMMA_found
49.         If COMMA in phrase:
50.             NEW_PHRASE = from start of sentence to COMMA position, excluding COMMA
51.             DO Procedure No_COMMA
52.             NEW_PHRASE = from start of sentence to V, ADD rest of sentence after COMMA
53.             DO Procedure COMMA_found

54.       End For
55. End
```

**Fig. 2.** Algorithm for Semantic Network Generation from natural language text (Source: author)

This research determined, through analysis of semantic network (SN) structures and language structure, that there was possibility of direct mapping of SPO to SVO, with unit of consideration being a sentence of phrase. Preliminary preprocessing of any text therefore determines the sentence or phrase lengths using full stops and commas, then processing that sentence or phrase (algorithm line 1 to 13). Further study on the sentence structure of SVO languages, such as Kiswahili, led to the realization that the key anchor to the SVO format of the language is the verb 'V'. The 'V' could be just one, but then it can have several subjects (S) and objects (O) lying on both of its sides.

The algorithm therefore determines the position of the verb (V) (line 14), then does several iterations to mine the POS on the left of the 'V', to get all the subjects. It then keeps them in a left hand side (LHS) array (line 22-23). The algorithm then mines the right side of the 'V' to determine all possible objects and stores them on the right-hand side (RHS) array (line 25-26). A final array of dimension LHS (L) x RHS (R) is now created. Any combination of L(1..n) + V + R(1..n) are candidate S-V-O triples, where 'n' is the total number of identified POS tagged subjects or objects (usually NOUNS). This generation of triples is what is achieved in line 38 of the algorithm.

For example, a sentence deconstructed into 2L subjects and 3R objects creates a 2x3 array with 6 possible SPO combinations. All these 6 combinations are based on only one single connector predicate being the 'V' (VERB). These 6 possibilities as SPO candidate triples are:
1. L1 + V + R1
2. L1 + V + R2
3. L1 + V + R3
4. L2 + V + R1
5. L2 + V + R2
6. L2 + V + R3

Where:
**Subject (S)**: L1, L2 are any subjects (usually a NOUN) – on the left hand side of the 'V'
**Object (O)**: R1, R2, R3 are also any objects (still usually NOUN) – on the right hand side of the 'V'
**Predicate (P)**: The static 'V' (VERB) is the predicate – in middle of the subject and object

There are therefore 6 potential SPOs triples of SVO format (NOUN subject + V + NOUN object), which is the expected structure of an SVO format language. That means that the algorithm is reformatting the SVO language into its basic SVO basic form.

However, not all the language constructs of the low-resource language shall necessarily be exactly in the SVO order. For example, some triples describe attributes such as 'is-a' relationship or even representation of attributes such as numbers or dates. The algorithm therefore creates a processing routine to look for such language constructs and then creates triples of 'is-a' type. This is done when a 'V' is not found in the sentence, while 'N' exists. Such additional rules can be quite many and are developed upon study of the POS tags of the NL text. These are implemented on lines 43 to 46 of the algorithm.

This algorithm therefore describes a rule-based system that leverages the POS tags of the language, then uses these basic rules to formulate candidate SVO triples and then augments the basic rules by any other inclusion or exclusion criteria to further increase or filter out any triples that should be added or excluded from the final SVOs for creating the SN.

All the final accepted SVOs are extracted and moved to a datastore in Resource Description Framework (RDF) format ready for further processing, including querying of the datastore using languages such as SPARQL. The triples stored as RDF can also be visualized on RDF graphs, which gives a visualization that the language has been decomposed into an interconnected structure. The proposed algorithm is then implemented using a typical programming language, such as Python, which is the chosen language to implement the algorithm in the tests done in this research as provided in the results section.

## 3.0 Results

TyDiQA dataset (Clark et al., 2020) is used as the data source for the following illustrations on how the algorithm processes NL text. The Swahili language portion of TyDiQA has a gold standard set in JavaScript Object Notation (JSON) file format. This file has 498 examples, with a context, a question, and an answer. Though TyDiQA is structured for machine learning tasks, we use it in the illustrations due to its ease of access and testing. It can also be used as a comparator with other machine learning methods since it is already formatted for machine learning. The full TyDiQA set is a collection of data in 11 languages i.e. Arabic, Bengali, English, Finnish, Indonesian, Kiswahili, Russian. Japanese, Korean, Thai and Telugu. And even in this set of eleven languages, Kiswahili is among the bottom three in terms of number of examples available (Wu & Wu, n.d.).

We show how the algorithm processes raw text from TyDiQA item ref. swahili--3141018404948436558-0 as an example. The JSON file content is reproduced below.

*{"title": "Chelsea F.C.", "paragraphs": [{"qas": [{"question": "Klabu ya Soka ya Chelsea ilianzishwa mwaka upi?", "answers": [{"text": "1905", "answer_start": 135}], "id": "swahili--3141018404948436558-0"}], "context": "Chelsea Football Club ni klabu ya mpira wa miguu ya nchini Uingereza iliyo na maskani yake Fulham, London. Klabu hii ilianzishwa mwaka 1905, na kwa miaka mingi sana imekuwa ikishiriki ligi kuu ya Uingereza. Uwanja wao wa nyumbani ni Stamford Bridge ambao una uwezo wa kuingiza watazamaji 41,837, wameutumia uwanja huu tangu klabu ilivyoanzishwa."}]}*

Like all items on TyDiQA, this is a Wikipedia article in the Swahili language. Wikipedia has an English language equivalent of this article that would provide the English meaning of the Kiswahili text (Contributors to Wikimedia projects, 2021). Though not the direct translation, the English version of Wikipedia is shown below:

*Chelsea Football Club is an English professional football club based in Fulham, West London. Founded in 1905, the club competes in the Premier League, the top division of English football. Chelsea is among England's most successful clubs, having won over thirty competitive honours, including six league titles and eight European trophies. Their home ground is Stamford Bridge.* (*Note that the number of spectators, 41,837 is not mentioned on the introductory paragraph of the English version of Wikipedia)

The algorithm creates a semantic network from the above NL text as per the flowchart in Fig. 1 and the description provided in section 2 above. This is done through the following stages:

### 3.1    Stage 1
Count the number of sentences to determine the number of iterations in the processing, in this case 3 sentences, which means 3 iterations through the SVO processing algorithm.

### 3.2 Stage 2

Undertake iterations as per sentence count from stage 1 as per the following process:

### 3.2.1 Iteration 1 – sentence 1:

Check for any COMMAs in the sentence and count them, in this case one comma. The rules of the algorithm assume that the comma is the enumeration type. Using the comma procedure of the algorithm, the sentence is decomposed into 2 parts as below:

***Phrase 1*** (everything upto the comma) - *Chelsea Football Club ni klabu ya mpira wa miguu ya nchini Uingereza iliyo na maskani yake Fulham*

***Phrase 2*** (everything upto the V, then everything after the comma) - *Chelsea Football Club ni klabu ya mpira wa miguu ya nchini Uingereza iliyo na maskani yake London.*

These two atomic phrases are processed in the full algorithm, which starts by undertaking the part of speech tagging (POST) of the first sentence ready for processing. An online tool is available for simple POST tasks for Kiswahili (aflat, 2020). It accepts any Kiswahili text and generates the POS tag.

#### 3.2.1.1 Phrase 1 processing (using the no comma) procedure:

Check for any VERBs (algorithm line 17), and since found (**DEF-V:ni**) - verb with no inflection, then check both the left and the right of the VERB and create an array of all nouns on the LHS and those on the RHS. Thereafter, create all possible combinations of triples of S-V-O based on the anchor verb. This process, as per algorithm line 37, generates triples ref. T1, T2 and T3 as shown on Table 1 below. Since there are no other connector verbs in phrase 1, the rest of the triples from phrase 1 (ref. T4 to T9) are generated by considering other rules on line 43 of the algorithm such as 'is-a' rules.

The first column of the table (Ref.) points to the line number of the algorithm of Fig. 2 that is being processed, while the triple in column 2 refers to the formulated candidate triples possible from the two or three POS tags considered. Note that the POS is not lemmatized, though it is possible to generate the lemmas using other tools such as Treetagger (treetagger, 2020).

**Table 1** – Parts of Speech extracted by Algorithm as suitable SVO triples

| Ref. | Triple | POS considered | | | Generated Triples (Turtle format) |
|---|---|---|---|---|---|
| 17 | POS | **PROPNAME** | **DEF-V:ni** | **N** | @prefix : <http://testing.123> |
| | T1 | *Chelsea* | *ni* | *klabu* | :chelsea :ni :klabu . |
| | T2 | *Football* | *ni* | *klabu* | :football :ni :klabu . |
| | T2 | *Club* | *ni* | *klabu* | :club :ni :klabu . |
| 43 | POS | **N** | **GEN-CON** | **N** | |
| | T4 | *klabu* | *ya* | *mpira* | :klabu :ya :mpira . |
| | T5 | *mpira* | *wa* | *miguu* | :mpira :wa :miguu . |
| | T6 | *miguu* | *ya* | *nchini* | :miguu :ya :nchini . |
| 43 | POS | **N** | **PROPNAME** | | |
| | T7 | *nchini* | *Uingereza* | | :nchi :ni :uingereza . |
| 43 | POS | **PROPNAME** | **CC** | **N** | |
| | T8 | *Uingereza* | *na* | *maskani* | :uingereza :ni :maskani . |

| 43 | POS | **N** | **PROPNAME** | | |
| | T9 | *maskani* | *Fulham* | | :maskani :ni :fulham . |
| 43 | POS | **N** | **PROPNAME** | | |
| | T10 | *maskani* | *London* | | :maskani :ni :london . |

**3.2.1.2 Phrase 2 processing (using the no comma) procedure:**
This is treated just like phrase 1, hence we repeat all the considerations done in processing phrase 1. Since phrase 2 is exactly same as phrase 1 apart from the last word, it generates all those triples already generated in phrase 1, apart from the last one which is different (T10 is generated instead of T9). The list of all POS considered for SVO suitability and other POS considered by other rules in generating semantic network triples is shown on columns 3 to 5 of Table 1 above.

**3.2.2  Iteration 2 – sentence 2:**
Follows the same process as above for sentence 1, by checking commas (one in this case) and the comma intention (conjunction to phrase 1). It then processes each phrase through the full algorithm.

**3.2.3  Iteration 3 – sentence 3:**
Follow the same process as above for sentence 1, by checking commas (one in this case) and the intention of the comma (conjunction to phrase 1). It then processes each phrase through the full algorithm. Note that the number 41,837 also has a comma, but the algorithm rules of number processing shall check for such to remove the comma for the number and only leave the one comma that combines the 2 phrases.

**3.3    Generating the Semantic Network**
All triples collected from the 3 sentences (3 iterations) are the triples that are then stored in a triple store, such as in RDF format. The triples are now ready for practical NLP tasks such as question answering. The diagrammatic visualization of the semantic network is shown in Fig. 3 below. The figure is generated from an online virtualization tool that accepts Turtle formatted RDF entries and then generate the graphical visualization (RDF Grapher, 2021). However, programming languages such as Python can also generate such graphs using existing libraries.

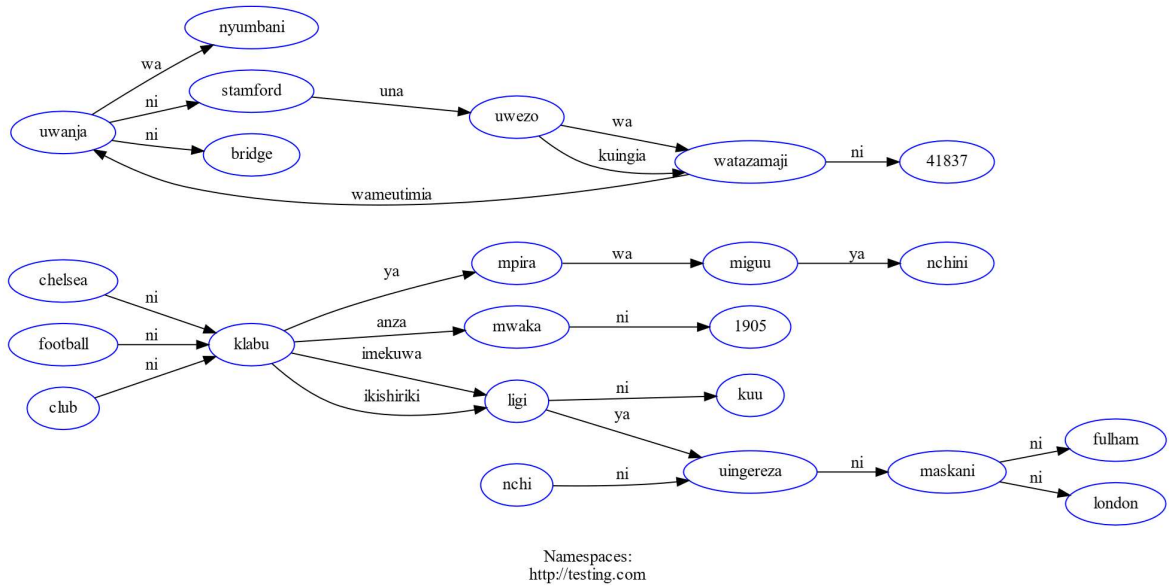

**Fig. 3.** Visualization of RDF triples created using the algorithm (Source: author)

### 3.4   Proof of Concept Question

We test the SN on a typical NLP QA task. The network in Fig. 3 is generated from TyDiQA text ID swahili--3141018404948436558-0, which was annotated with a question as below:
*Q-Klabu ya Soka ya Chelsea ilianzishwa mwaka upi?*
(Q-When was Chelsea Football Club founded?)

This answer is easily seen as the relationship between 'Chelsea' and 'founded' as per the visualization on Fig. 3, though we query the datastore using SPARQL shown below (using Python RDFLIB module), since the triples are stored as RDF.

*SELECT ?s ?p ?o*
   *WHERE {*
   *{:chelsea ?p ?o .}*
   *UNION*
   *{?s ?p :mwaka .}*
   *UNION*
   *{:mwaka ?p ?o .}*
   *}*

Note that in the Kiswahili language text, the concepts (club name and year founded) are in different sentences, however, the nodes and edges of the SN interconnect the whole story to show the inter-relatedness of the domain. Additionally, words in both the texts of the context and the text of the question would usually need to be lemmatized so that the lemmas of the question text is aligned to the lemmas of the story text e.g. *anza/anzishwa* (found/founded) need to point to the same concept in the SN.

The response obtained is shown below, with predicates and objects indicated in bold:

*http://testing.123/**ni** http://testing.123/**klabu***
*http://testing.123/**klabu** http://testing.123/**anza***
*http://testing.123/**ni** http://testing.123/**1905***

With the final triple giving the figure of '1905', which is the correct answer for this question. Note that our query did not exploit the concept of date extraction, but rather an object extraction, with the object being the year/date. It could have been possible to create RDF triples that indicate that '1905' is a date, then create a query to look for a date object on the graph.

### 3.5    Testing the Algorithm on different QA datasets

This algorithm was tested on 3 different question-answer datasets. The first was the Tusome corpus (Kenya Ministry of Education, n.d.; Piper et al., 2018), which is a collection of texts for early childhood education in Kenya. The exact match score for the sampled 33 QAs was 63.7% i.e. 21 correctly answered. The second dataset was the TyDiQA set (Clark et al., 2020). In TyDiQA, the low-resource language of Swahili has 17,613 training examples, 2,288 development set examples and 2,278 in the test set. The gold standard set that supports SQuAD version 1.1 is a JavaScript Object Notation (JSON) file of 498 examples, with context, question, and answer. The research tested the data on this gold standard set.

We used a purposive sample set of 54 questions to test the different question types. We dynamically created individual SNs of each story texts, then subjected these SNs to QA task on the fly. The results were 35/54 correctly answered questions (64.8% exact match accuracy). The distribution of questions tested in this set is shown in Table 2 below.

**Table 2** – Analysis of Questions subjected to the SN generated from texts in the TyDiQA corpus

| Question Type | Date/Yr | Num | What | Define | Where | Which | Who | Total |
|---|---|---|---|---|---|---|---|---|
| Total | 18 | 5 | 5 | 3 | 12 | 8 | 3 | 54 |
| Correct (EM) | 12 | 3 | 4 | 0 | 8 | 6 | 2 | 35 |

The final QA dataset used to test the algorithm was the KenSwQuAD dataset (B. W. Wanjawa et al., 2023). This is a QA dataset of 7526 QA pairs based on 2585 texts. The KenSwQuAD is a gold standard set Swahili QA dataset for testing QA systems. We sampled 365 questions using purposive sampling to ensure that each question type is selected when picking the texts to create the SNs in readiness for the QA task. Table 3 below shows the distribution of question types and the QA results. We observe that 287/365 QAs were correctly answered, hence 78.6% exact match performance.

**Table 3** – Analysis of Questions subjected to the SN generated from texts in the KenSwQuAD corpus

| Question Type | Date/Yr | Num | What | Define, How, Why | Where | Which | Who | Total |
|---|---|---|---|---|---|---|---|---|
| Total | 15 | 15 | 125 | 36 | 25 | 76 | 73 | 365 |
| Correct (EM) | 8 | 9 | 118 | 0 | 21 | 66 | 65 | 287 |

## 4.0 Discussions

The research formulated an algorithm that is useful in creating structure out of a natural language (NL) text of a low-resource language such as Kiswahili. Structuring language, or data, is usually a prerequisite step before further processing or mining of information. For NL, structuring has been done using methods such as term-frequency-inverse-document frequency (TF-IDF), word embeddings and deep learning that include transformer models, depending on the ultimate use of the structured language.

Many of the existing machine learning models which are used to structure NL need training data, which is readily available for high resource languages e.g. English, French, Chinese, Arabic etc. However, low-resource languages, such as Kiswahili and many other African languages how few datasets for use in machine learning models. Despite this drawback, languages, including low-resource, are important for human communication and are useful in disseminating information that can be lifesaving, such as during disasters, natural calamities, medical emergencies e.g. during the Corona Virus Disease (COVID) pandemic and even terrorism incidences.

However, some knowledge representation systems such as semantic networks (SNs) do not necessarily need training data. These methods have traditionally been applied in other domains that deal with unstructured data processing to get inter-relatedness of entities or objects e.g. Facebook (X. Li & Boucher, 2013), LinkedIn Graph (Markovic & Nelamangala, 2017) and Google Knowledge graph (Singhal, 2012).

The same reasoning of inter-relatedness of objects has been used in this research to structure natural language text. Some low-resource languages, such as Kiswahili, have a generic structure described as subject-verb-object (SVO) – a general inter-relatedness of a triple of subject-verb-object. On the other hand, semantic networks have a triple of subject-predicate-object (SPO) relationship, hence an SVO-to-SPO mapping is possible.

Mapping SVO pattern of a language to SPO structure of a semantic network is however not trivial, since language sentences are hardly a three-word triple of S then V then O. The research algorithm provided the method of identifying and extracting suitable SVO from NL text to map into the SPO of SNs. The method is based on rules that pick out these SVO triples. Only the part of speech (POS) is required, not the words of the language itself. The main consideration is to start by identifying the key verb (V) as the predicate, then work both backwards and forward to determine the subject(S) and object(O), usually nouns. Finally, increase coverage by considering other language constructs e.g. 'is-a' relationships. Ultimately, there cannot be a ruleset that covers all aspects of the very expressive nature of NL. Some sentence constructs shall still be parsed incorrectly or not at all.

Since only the POS tags are necessary in creating the SN, any language whose POS is known can therefore be structured into a SN. This research only tested the direct link between SVO language structure to the SPO structure of the SN. However, by similar reasoning, an algorithm can still transform even an SOV language to a SPO structure of a SN.

The results got for the proof of concept task of QA from the SNs created from NL test realized an accuracy of 64.8% on exact match for the sampled TyDiQA set (Clark et al., 2020) and 78.6% exact match on the KenSwQuAD dataset (B. W. Wanjawa et al., 2023). These results show that the NL texts have been structured into SNs that now represent these domains, hence are capable of practical tasks on the NL text, with no training data. The only prerequisite is the POS tagging of the NL text.

However, we observe that this SN based method performs quite well at object-enquiry type of questions (who, what, where, when/date) and not on the explanatory types of questions (define, why). This is expected, since SN by its very nature is an object linking method, where objects such as language POS tags are interlinked using connectors, which are also POS tags. Querying is therefore an enquiry into the objects within the SN. Reasoning questions require much more than simple object linking within a context. As reported on that KenSwQuAD research, even deep learning methods could not perform well on QA task based on the same dataset due to the limited training data, achieving only an F1 score of 59.4% and exact match of 48% (B. W. Wanjawa et al., 2023).

Some shortcomings and challenges with the use of the proposed rule-based algorithm are noted when processing typical language POS constructs. One problem noted is the determination of intention of the COMMA part of speech. In some cases, the comma signifies the enumeration of related items in a list, while in other cases it may signify the start of a new phrase that is related, or not, to the first phrase. Commas can also be used after initial startup words or to split phrases, or mark pauses, and even within numeric values. The same problem of interpretation is observed with semicolons, colons, slashes, hyphens, and apostrophes. The algorithm has just modelled a few commonly used constructs and usually defaults to skipping any other possible extractions that are not on the ruleset. Processing direct speech or drama/plays is still a problem with rule-based or other methods of NL processing.

Other language processing challenges in general include named entity recognition (NER) that should be done at POS tagging. Currently, without NER tool, the named entity is just decomposed into separate nouns, and the connecting verb (V) is just likely to link each left-hand side noun with some right-hand side object. This is manifested in the example where the named entity 'Chelsea Football Club' is decomposed into three separate proper nouns, instead of one single named entity. Each of the proper nouns is then equated to be a club, instead of one single named entity being the club.

Coreference resolution is another challenge that we face when processing low resource language text. In our three-sentence example of Fig. 3, we see a graph that is not fully connected, despite this being one connected domain. The lack of connectedness is caused by how sentence 3 is processed, since it no longer mentions the subject by name, but uses 'it' instead. An NLP tool to resolve this article 'it' into the proper name of the subject is not available for now. However, these shortcomings also present an opportunity for further research into resolving these identified challenges.

### 5.0 Conclusion

This research formulated an algorithm for use in transforming a natural language text of a low-resource language into a semantic network. The low-resource language tested in this research was an SVO-type language of Kiswahili. Tests done on the Kiswahili language texts structured into semantic networks using the algorithm have confirmed that the language is given a structure. The task of question answering has further been used as proof of concept confirming that the structured language is capable of practical use including querying using existing querying languages.

Structuring the natural language text into triples that fit the semantic network structure is however not trivial and hence a guiding algorithm is necessary. This is because the natural language text of Kiswahili, though described as SVO-type, does not necessarily mean that the sentences are a series of S then V then O. It is much more complex, and an algorithm with its ruleset decomposes the complexity to guide on the exact processing steps in generating the final SVO triples. The research recognized that anchoring on the V(verb) as the starting point, it is then possible to process the suitable subjects and objects, and then combine the suitable SVO triples for extraction into a datastore.

The algorithm has been tested in practical datasets of Tusome (Kenya Ministry of Education, n.d.), TyDiQA (Clark et al., 2020) and KenSwQuAD (B. W. Wanjawa et al., 2023), where we confirm that it is possible to create semantic networks by just leveraging on the POS tags of a language, achieving 78.6% exact match performance on the KenSwQuAD dataset, with no training data. Only the parts of speech tagging were the prerequisite. This structured language is then of practical computer information processing tasks such as QA. Alternative methods such as machine learning, deep learning and even transformer models, that are highly successful in other high-resource languages, are unfortunately not applicable in the case of many of the low-resource languages that do not have training data.

Challenges still abound, such as dealing with direct speech within the texts and the inability to create rules that cover all language constructs e.g. intentions of punctuation such as commas, semicolons, and hyphens. Additional challenges include dealing with named entities which need to be resolved into noun objects. Co-reference resolution also stands in the way of linking language concepts into a jointed network during processing. Other non-SVO relationships are also difficult to catch and model e.g. 'is-a' relationships. An increased rule set may assist in resolving some shortcomings, though natural language is so expressive hence difficult to fully model. As was done with the high-source language of English, deliberate research and development is needed for low resource languages to come up with all these tools that assist in resolving these challenges.

There is also the challenge of being able to perform the initial part of speech (POS) tagging itself, which is a prerequisite for the algorithm. This however is also an opportunity to focus research interests on POS tagging as a priority, to in turn start exposing these languages to machine processing. Language dictionaries, which are usually available in some form, can be a starting point in generating POS tags. Only the POS tags and the application of an algorithm such as what we propose in this research, are needed to then create a knowledgebase of the language ready for practical use.

Researchers still need to work towards collection of the datasets of low-resource languages so that existing methods that perform well in high-resource languages, such as deep learning models, can also be tried for low-resource languages to see whether there is improved performance compared to the rule-based systems. An immediate research opportunity is using the algorithm to process SOV-type low-resource languages. A tweak on the algorithm should be able to structure such SOV language into SPO structure of a semantic network. This can be done after the study of the SOV sentences and an analysis of how meaning is formed from the texts of these languages by SOV linkages. This should further extend the coverage of the potential languages for structuring based on this research algorithm.

**Funding**
This research did not receive any specific grant from funding agencies in the public, commercial, or not-for-profit sectors.

**Conflicts of interest**: none